\documentclass[lettersize,journal]{IEEEtran}
\usepackage{amsmath,amsfonts}
\usepackage{algorithmic}
\usepackage{algorithm}
\usepackage{booktabs}
\usepackage{inputenc}
\usepackage{fontenc}
\usepackage{multirow}
\usepackage{array}
\usepackage{graphicx}
\usepackage{pdflscape}
\usepackage[caption=false,font=normalsize,labelfont=sf,textfont=sf]{subfig}
\usepackage{lettrine}
\usepackage{textcomp}
\usepackage{rotating}
\usepackage{stfloats}
\usepackage{url}
\usepackage{verbatim}
\usepackage{footmisc} 

\makeatletter
\renewcommand\@makefntext[1]{%
    \hspace*{1em}#1
}
\makeatother
\usepackage[dvipsnames]{xcolor} 
\usepackage{cite}

\usepackage[colorlinks=true,linkcolor=blue,citecolor=blue,urlcolor=blue]{hyperref}
\usepackage{titlecaps} 
\let\clearpage\relax

\begin{document}

\title{Dendritic Convolution for Noise Image Recognition}

\author{Jiarui~Xue\textsuperscript{\dag}, Dongjian~Yang\textsuperscript{\dag}, Ye~Sun, and Gang~Liu*}
\footnotetext{*Corresponding authors: Gang Liu(gangliu\_@zzu.edu.cn)}
\footnotetext{This work is supported by the National Natural Science Foundation of China (62303423), the STI 2030-Major
Project(2022ZD0208500), Postdoctoral Science Foundation of China
(2024T170844,2023M733245), the Henan Province key research
and development and promotion of special projects (242102311239),
Shaanxi Province key research and development plan(2023GXLH-
012)}
\footnotetext{Jiarui Xue and Dongjian Yang contributed equally to this work.}

\maketitle
\begin{abstract}
In real-world scenarios of image recognition, there exists substantial noise interference. Existing works primarily focus on methods such as adjusting networks or training strategies to address noisy image recognition, and the anti-noise performance has reached a bottleneck. However, little is known about the exploration of anti-interference solutions from a neuronal perspective.This paper proposes an anti-noise neuronal convolution. This convolution mimics the dendritic structure of neurons, integrates the neighborhood interaction computation logic of dendrites into the underlying design of convolutional operations, and simulates the XOR logic preprocessing function of biological dendrites through nonlinear interactions between input features, thereby fundamentally reconstructing the mathematical paradigm of feature extraction. Unlike traditional convolution where noise directly interferes with feature extraction and exerts a significant impact, DDC mitigates the influence of noise by focusing on the interaction of neighborhood information. Experimental results demonstrate that in image classification tasks (using YOLOv11-cls, VGG16, and EfficientNet-B0) and object detection tasks (using YOLOv11, YOLOv8, and YOLOv5), after replacing traditional convolution with the dendritic convolution, the accuracy of the EfficientNet-B0 model on noisy datasets is relatively improved by 11.23\%, and the mean Average Precision (mAP) of YOLOv8 is increased by 19.80\%. The consistency between the computation method of this convolution and the dendrites of biological neurons enables it to perform significantly better than traditional convolution in complex noisy environments.
\end{abstract}

\begin{IEEEkeywords}
Anti-interference, Convolution, Dendritic structure, Image classification, Noise robustness, Object detection.
\end{IEEEkeywords}

\section{Introduction}
\IEEEPARstart{I}{n} real-world image recognition scenarios, there is often substantial noise interference\cite{ref1} \cite{ref2} \cite{ref3}. For instance, Gaussian noise commonly generated during the operation of image sensors \cite{ref4}; Poisson noise particularly prominent in low-light imaging \cite{ref5}; salt-and-pepper noise caused by signal pulse interference \cite{ref6}; and speckle noise, which is prevalent in medical ultrasound and remote sensing imaging and formed by coherent interference induced by medium scattering, among others \cite{ref7} \cite{ref8}. These types of noise are widely present in practical imaging systems and pose a persistent challenge to the robustness of models. 
\begin{figure*}[t!]  
\centering
\includegraphics[width=1\linewidth]{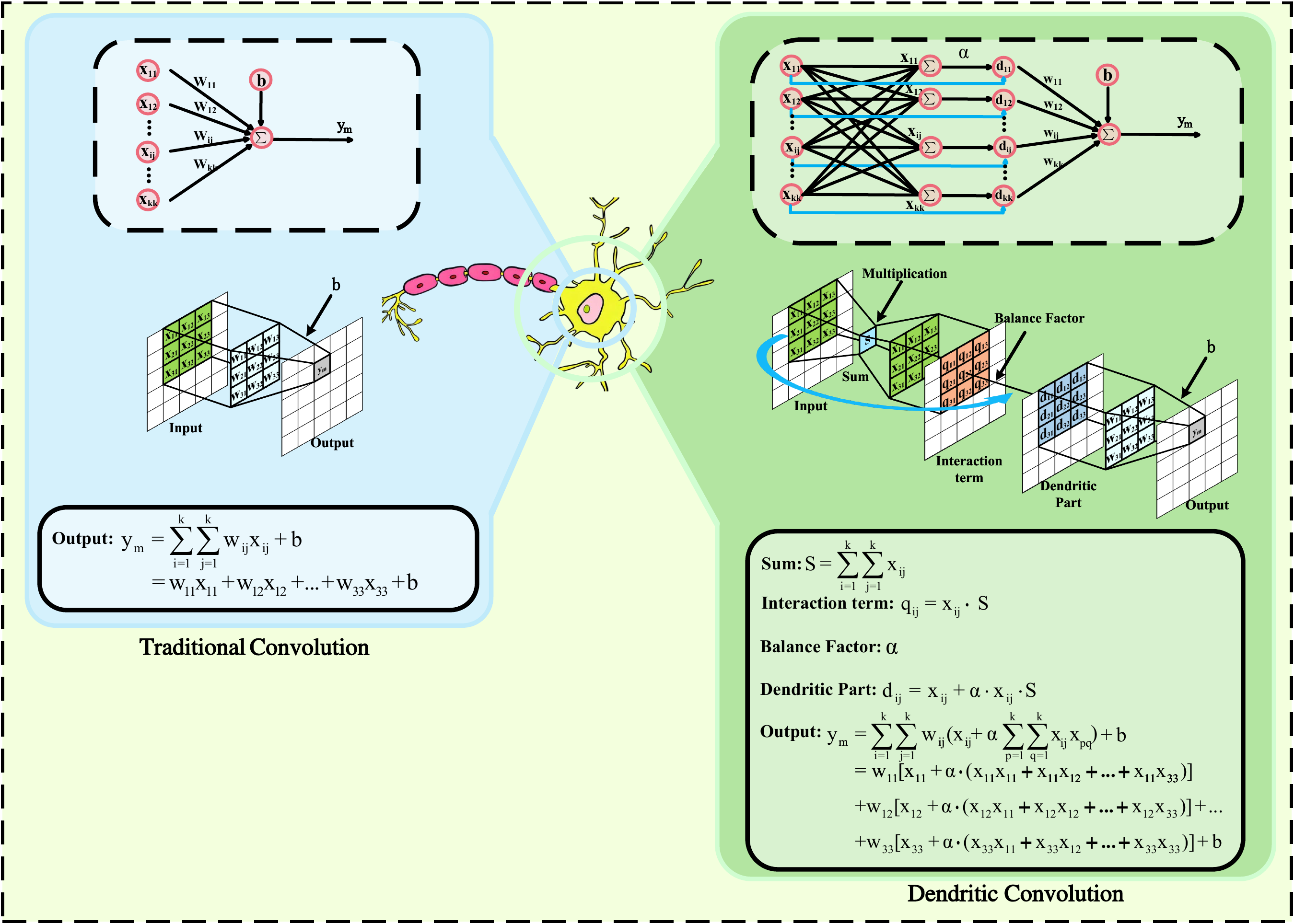}  
\caption{Schematic Diagram of Neuron Simulation Comparison Between Traditional Convolution and Dendritic Convolution}
\label{fig:comparison}
\end{figure*}

To address this, researchers have proposed a variety of anti-interference methods to enhance the robustness of models in noisy environments. These efforts are primarily reflected in two aspects: architectural improvement and data augmentation. At the architectural level, more efficient modules or mechanisms are introduced to strengthen the representational capacity of networks. A case in point is the widespread adoption of the ResNet \cite{ref9}. It effectively alleviates the gradient vanishing problem in deep networks through residual connections, laying a foundation for building deeper and more powerful backbone networks, ensuring the stability of feature extraction, and indirectly boosting model robustness. To further tackle complex noise interference, the mainstream trend is to incorporate attention mechanisms. For example, YOLOv4 \cite{ref10} and its subsequent versions attempt to integrate mechanisms such as Squeeze-and-Excitation (SE) attention \cite{ref11} or Convolutional Block Attention Module (CBAM) \cite{ref12}; by calibrating channel or spatial feature responses, these mechanisms enable the model to focus on key information.At the data level, data augmentation strategies \cite{ref13} \cite{ref14} have become a common approach to improve the generalization ability and robustness of models. Similarly, modern frameworks like YOLOv4 have built-in Mosaic data augmentation [10], which aims to proactively enable the model to learn samples in complex environments during training. Besides, adversarial training is also frequently used to enhance the noise robustness of models. Goodfellow et al. have experimentally demonstrated that adversarial training can enhance the model’s resistance to directional noise such as adversarial tiny perturbations \cite{ref15}. Although with the iteration of architecture versions, most of the aforementioned anti-interference methods have been integrated into current mainstream image recognition architectures and object detection frameworks, they essentially add extra modules on top of networks built with traditional convolution or perform transformational enhancement on input data before feeding it into the network. These methods are difficult to fundamentally address the inherent limitations of traditional convolution. Specifically, its linear weighted summation mode cannot well simulate the nonlinear integration and preprocessing functions of dendrites in biological neurons for multiple input signals, which results in models being unable to adaptively filter out noise information like biological neural systems \cite{ref16} \cite{ref17} \cite{ref18} \cite{ref19}. Therefore, addressing this issue from the perspective of biological neurons is a crucial direction to respond to the current scarcity of research and enhance the noise resistance of models.

To address this situation, this paper proposes and designs a novel convolution called Dendritic Convolution (DDC). It is built upon the finding of the Dendrite Net (DD) proposed by our team  \cite{ref20}. DD is a white-box algorithm that simulates biological dendritic computation. It realizes nonlinear interactions between input features via Hadamard product, achieves the XOR logic preprocessing function for local features, and simulates the biological functions of dendrites including signal reception, logical preprocessing and information transmission, which provides underlying computational paradigm support for the subsequent design of Dendritic Convolution. Unlike traditional convolution, DDC introduces a dendrite-like nonlinear local integration mechanism. After extracting local regions and reorganizing neighborhood features, it draws on the Hadamard product of DD to simulate the integration of multi-synaptic signals by dendrites. The generated features can fully reflect local feature interactions and adaptively capture adjacent correlations (see Fig. 1). Meanwhile, as a plug-and-play module, DDC retains the parameter design of traditional convolution, can directly replace traditional convolution, and can be integrated into existing frameworks without modifying the network structure.

From the experimental results of image classification and object detection, the noise robustness of image classification ( YOLOv11-cls, VGG16 \cite{ref21} and EfficientNet-B0 \cite{ref22}) is improved by 6.25\%--11.23\%. The noise robustness of object detection ( YOLOv11, YOLOv8 and YOLOv5) is improved by 10.69\%--19.80\%, and the effect is shown in TABLE I. The \( A_{avg} \) calculation fornmula is as follows:
\begin{equation}
\label{deqn_ex1a}
A_{avg}=\frac{1}{6} \sum_{i=1}^{6} A_{i}
\end{equation}

Where \( A_i \) corresponds to the accuracy under Gaussian, Poisson, Salt\&Pepper, Speckle, Rayleigh and Gamma noise, respectively.The contributions of this paper are as follows:
\begin{table*}
\centering
\caption{Relative Improvement Percentage Of Model Accuracy}
\label{tab:accuracy_improvement}
\resizebox{\linewidth}{!}{%
\begin{tabular}{lcccccccc}
\toprule
\textbf{Task} & \textbf{Model} & \textbf{Gaussian} & \textbf{Poisson} & \textbf{Salt\&Pepper} & \textbf{Speckle} & \textbf{Rayleigh} & \textbf{Gamma} & \textbf{Average} \\
\midrule
\multirow{3}{*}{Classification} 
& YOLOv11-cls & 0.64\% & 11.20\% & 1.08\% & -0.03\% & 10.64\% & 13.96\% & 6.25\% \\
& EfficientNet-B0 & 13.36\% & 11.47\% & 11.75\% & 14.55\% & 10.31\% & 5.91\% & 11.23\% \\
& VGG16 & 5.93\% & 22.54\% & 13.66\% & 5.29\% & 6.58\% & 1.80\% & 9.30\% \\
\midrule
\multirow{3}{*}{Detection}
& YOLOv11 & 4.08\% & 21.25\% & 13.19\% & 3.44\% & 13.30\% & 17.12\% & 12.06\% \\
& YOLOv8 & 6.85\% & 34.17\% & 18.36\% & 5.99\% & 20.42\% & 33.00\% & 19.80\% \\
& YOLOv5 & -1.12\% & 9.01\% & 11.50\% & -0.37\% & 10.00\% & 35.10\% & 10.69\% \\
\bottomrule
\end{tabular}%
}
\end{table*}
\begin{enumerate}
    \item In this paper, a convolution DDC that simulates biological dendritic computing logic is proposed for the first time. Compared with the traditional convolution, the convolution is similar to the biological dendrites, can resist various types of noise, and can replace the traditional convolution to achieve plug-and-play functions.
    
    \item In this paper, for image classification and object detection tasks, the proposed DDC is substituted into multiple models. The performance of the original models is significantly improved, with the maximum improvement reaching 19.80\%.
\end{enumerate}

\section{Related Work}
\subsection{Convolution}
The success of Convolutional Neural Networks (CNNs) lies in their ability to hierarchically extract local features \cite{ref23} \cite{ref24}, yet the linear weighted summation mechanism they rely on at the bottom layer has inherent flaws in complex interference environments. Traditional convolution performs linear combination of features in local regions; although this paradigm can capture basic features such as edges and textures, it is difficult to simulate the adaptive noise filtering capability of the biological nervous system. Researchers have carried out extensive explorations around the optimization of convolutional structures. For example, in terms of improving model robustness, Deformable Convolution proposed by Dai et al \cite{ref25}. adds learnable 2D offsets to the regular sampling grid of standard convolution, enabling sampling positions to adapt to the geometric deformation of objects. It can handle occlusion and complex background interference without additional supervision, thus better adapting to changes in object morphology; Adaptive Rotated Convolution (ARConv) designed by Pu et al. \cite{ref26} predicts the rotation angles and combination weights of convolution kernels based on input features, allowing the convolution kernels to meet the feature extraction needs of rotated objects and also adapting to the feature extraction of multi-oriented objects in a single image; Linear Deformable Convolution (LDConv) \cite{ref27} generates initial sampling positions for convolution kernels of arbitrary sizes through a coordinate generation algorithm, and then adjusts the sampling shape in combination with offsets. It not only realizes irregular convolutional feature extraction but also makes the parameters grow linearly, balancing performance and hardware overhead.

Although the aforementioned methods can improve task adaptability in specific scenarios, none of them have reconstructed the convolutional operation from the bottom layer of neuron function simulation. These methods either adjust the morphology and parameters of convolution kernels or add feature processing modules; essentially, they only make modifications based on the traditional linear-weighted paradigm, making it difficult to fundamentally break through the inherent limitations of traditional convolution. In contrast, DDC reconstructs the underlying logic of convolutional operations. By simulating the multi-branch nonlinear integration characteristics of dendrites, it enables the convolutional operation to enhance effective information and suppress redundant signals at the initial stage of feature extraction. This bottom-layer design starting from the neuron level does not simply aim to solve the noise problem; instead, through a computational logic closer to that of biological systems, it endows the model with stronger feature extraction capabilities and adaptive capabilities, thereby providing more fundamental support for improving the robustness and accuracy of the model.
\subsection{Bio-Inspired Dendritic Models}
In biological neurons, dendrites do not simply transmit signals; instead, they perform nonlinear integration of multi-synaptic inputs through complex branching structures to achieve efficient information preprocessing and feature selection. This unique information processing method endows the biological nervous system with strong information processing capabilities and adaptability to complex signals. Inspired by this biological mechanism, exploring how to integrate the computational characteristics of dendrites into artificial models has become an increasingly valued research direction. For example, the dendritic Artificial Neural Network (dANN) \cite{ref28} simulates the sparse connection characteristics of biological dendrites, significantly reducing trainable parameters in image classification tasks while enhancing anti-overfitting ability, thus demonstrating the advantage of parameter efficiency. The DMobileNet model designed by Gao et al. \cite{ref29} combines the Dendritic Neuron Model (DNM) with the lightweight convolutional network MobileNet; in brain tumor detection tasks, it significantly improves the model’s ability to capture edge features and nonlinear expression ability, enhancing the nonlinear modeling capability of the classifier while maintaining low computational overhead. Tang et al. \cite{ref30} expanded the traditional dendritic neuron model into a Dendritic Neural Network (DNN); by introducing synaptic flexibility and a dropout mechanism, they improved the model’s ability to handle complex tasks and generalization performance, solving the limitation of a single dendritic neuron model in multi-classification tasks. The Dendrify framework developed by Pagkalos et al. \cite{ref31} integrates dendritic characteristics into spiking neural networks, successfully constructing a simplified compartmental neuron model with both biological accuracy and computational efficiency, providing efficient and bio-inspired tool support for neuromorphic computing.

Although existing dendritic models have successfully combined some characteristics of dendrites with artificial neural networks and made progress in multiple aspects, their integration approaches mostly focus on adjustments and designs at the network architecture level. Essentially, these are upper-layer improvements based on existing network frameworks. In contrast, the dendritic convolution proposed in this paper directly integrates the nonlinear integration and information preprocessing logic of biological dendrites into the underlying operational logic. Moreover, it reconstructs the computational rules of convolution, allowing the information processing characteristics of dendrites to take effect from the initial stage of feature extraction. This enables the model to naturally possess anti-interference and feature selection capabilities similar to those of biological dendrites during the feature acquisition process, achieving a more fundamental and in-depth integration of dendritic characteristics with artificial neural networks.
\section{Proposed Method}
\subsection{Traditional Convolution}
The traditional convolution operation simulates the linear weighted summation mechanism of neuron cell body to realize the feature extraction of local receptive field. The mathematical model can be expressed as :
\begin{equation}
{y} = \sum_{i=1}^{k} \sum_{j=1}^{k} {w}_{ij} {x}_{ij} + {b}
\end{equation}
Where \({w}_{ij}\)\ denotes a learnable weight matrix of size \(k \times k\), \({x}_{ij}\)\ represents the input feature elements within the local receptive field, and \({b}\)\ is the bias term. Its structural diagram is shown in Fig. 2 .The traditional convolution operation performs linear weighted summation between the learnable weight kernel and the input feature map to simulate the function of the neuron soma. Due to the lack of capability in handling nonlinear interactions of neighboring features in this base structure, it is prone to loss of effective information and degradation of model performance under noise interference.
\subsection{The Proposed Dendritic Convolution}
\subsubsection{Dendritic Convolution Module}
\begin{figure*}[t!]  
\centering
\includegraphics[width=1\linewidth]{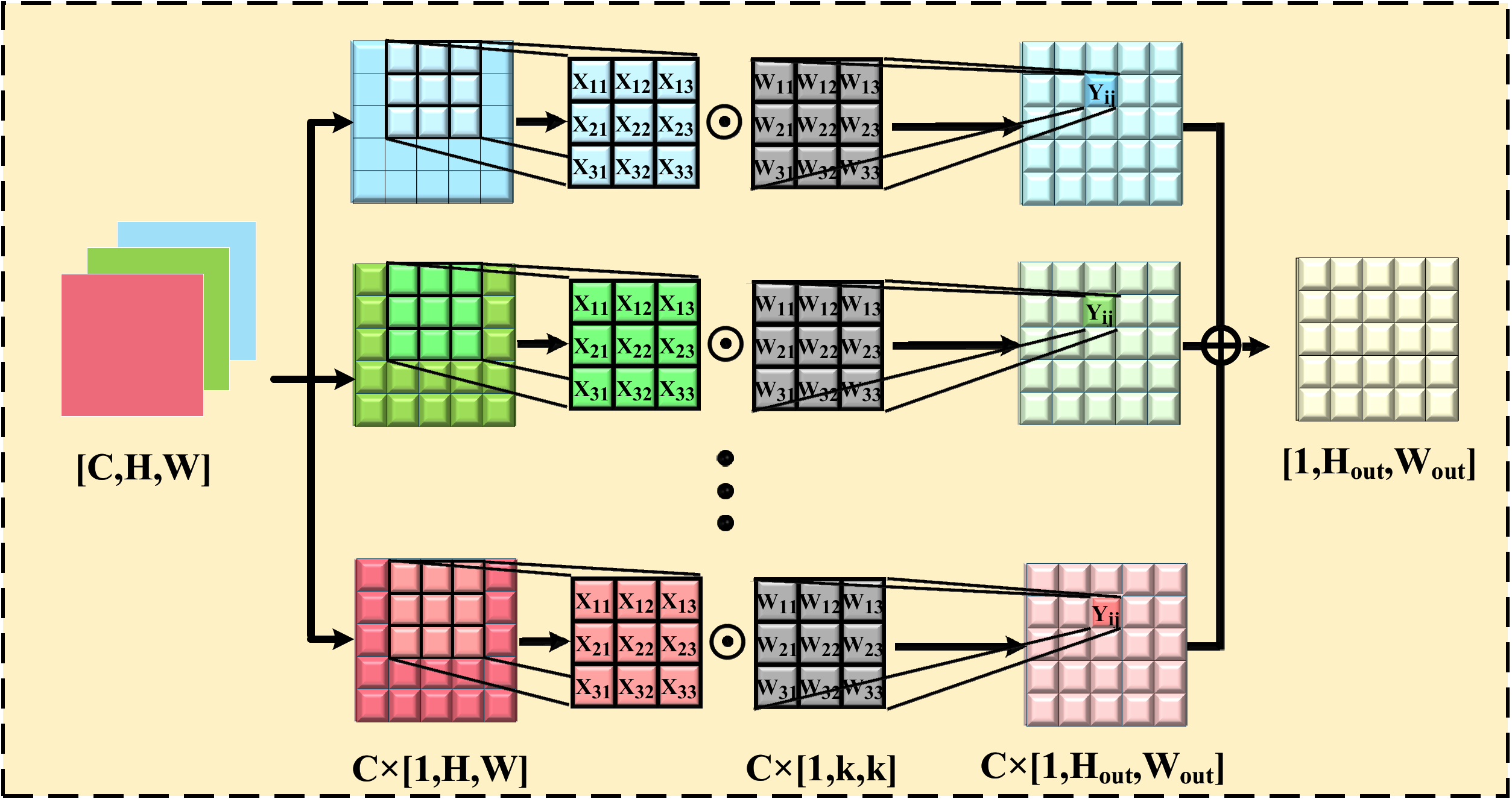}  
\caption{Schematic diagram of Traditional Convolution operation. }
\label{fig:comparison}
\end{figure*}
\begin{figure*}[t!]  
\centering
\includegraphics[width=1\linewidth]{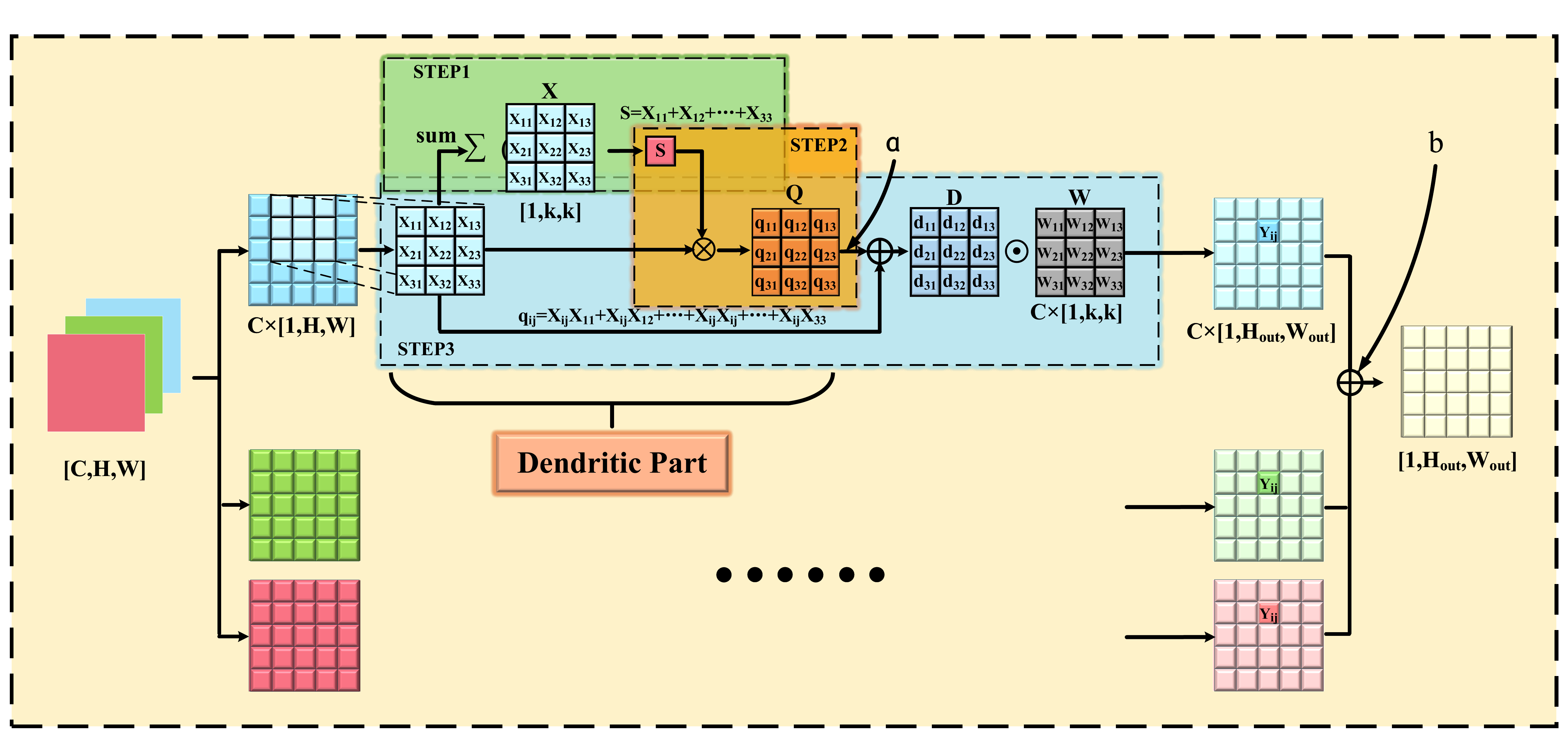}  
\caption{Schematic diagram of DDC feature extraction.}
\label{fig:comparison}
\end{figure*}
Inspired by the signal processing mechanism of biological dendrites, this paper proposes a Dendritic Convolution module that integrates local feature interaction and global semantic information. Its innovation lies in simulating the nonlinear integration process of dendritic branches through quadratic interaction terms, thereby achieving adaptive capture of global correlations. The module operation is defined as follows: 
\begin{equation}
{y} = \sum_{i=1}^{k} \sum_{j=1}^{k} {w}_{ij} \left( {x}_{ij} + \boldsymbol{\alpha} \sum_{p=1}^{k} \sum_{q=1}^{k} {x}_{ij} {x}_{pq} \right) + {b}
\end{equation}

Where \({x}_{ij}\) represents each element of the input feature block, simulating the main trunk of a neuron; \(\sum_{p=1}^{k} \sum_{q=1}^{k} {x}_{ij} {x}_{pq}\) is a quadratic interaction term, simulating the interaction information received by dendritic branches; \(\boldsymbol \alpha\) is a balancing factor, simulating the balancing effect in dendritic signal integration and balancing the signal integration intensity; \({w}_{ij}\) is each element in the learnable weight matrix, simulating the weight redistribution of dendrites to synapses at different positions. This design constitutes a dendritic residual structure, where the main trunk in the dendritic neural network is responsible for maintaining the stable transmission of original information, while the branches introduce enhanced information, thereby avoiding information loss during transmission. The structure not only enhances the richness of information but also enables the model to better capture and utilize global correlation. Its schematic diagram is shown in Fig. 3.
\subsubsection{Feature Interaction Mechanism}

This mechanism, by simulating the way dendrites process signals in biological neurons, introduces a feature containing global interaction information between features on the basis of traditional convolution operations, thereby enhancing the model’s representation capability. The implementation of this process can be divided into the following three steps.
\paragraph{Global Information Aggregation}
First, a local feature block \(\mathbf{X}\) of size \(k \times k\) is extracted from the input features (its matrix representation is shown in Eq 4).
\begin{equation}
\mathbf{X} = (x_{ij}) = \left( \begin{array}{ccc}
x_{11} & \cdots & x_{1k} \\
\vdots & \ddots & \vdots \\
x_{k1} & \cdots & x_{kk}
\end{array} \right) \in \mathbb{R}^{k \times k}.
\end{equation}

This step aims to obtain global information of all elements within each local block. As shown in Fig. 3 (STEP1), all elements in the feature block \(\mathbf{X}\) are summed to obtain the global total \(\mathbf{S}\):
\begin{equation}
\mathbf{S} = \sum_{i=1}^{k} \sum_{j=1}^{k} x_{ij} = x_{11} + x_{12} + \ldots + x_{ij} + \ldots + x_{kk} 
\end{equation}
This operation aggregates the neighborhood interaction information within the local receptive field.

\paragraph{Interactive Feature Generation}

Next, to generate interactive features, the global total \(\mathbf{S}\) obtained in the previous step is subjected to scalar multiplication with the original feature block \(\mathbf{X}\) to obtain the interactive feature block \(\mathbf{Q}\), i.e., the dendritic part.
\begin{equation}
\mathbf{Q} = \mathbf{X} \cdot \mathbf{S} = (q_{ij}) = 
\begin{pmatrix}
q_{11} & \cdots & q_{1k} \\
\vdots & \ddots & \vdots \\
q_{k1} & \cdots & q_{kk}
\end{pmatrix} \in \mathbb{R}^{k \times k}.
\end{equation}

\begin{equation}
q_{ij} = {x}_{ij} \times \mathbf{S} = x_{ij}x_{11} + x_{ij}x_{12} + \ldots + x_{ij}x_{ij} + \ldots + x_{ij}x_{kk}
\end{equation}

As shown in Fig. 3 (STEP2), the significance of this operation is to make each element \({x}_{ij}\) in the feature block interact with the global information \(\mathbf{S}\). The result is a interactive feature block \(\mathbf{Q}\) that encodes the consistency relationship between elements and their neighborhoods.
\paragraph{Residual Fusion and Output}

Finally, adopting the idea of a residual structure, the generated interactive feature \(\mathbf{Q}\) is added to the original trunk feature \(\mathbf{X}\) to construct a feature representation that integrates global interaction information. As shown in Fig. 3 (STEP3), its output is:
\begin{equation}
\mathbf{D} = \mathbf{X} + \boldsymbol{\alpha} \times \mathbf{Q} = (d_{ij}) = 
\begin{pmatrix}
d_{11} & \cdots & d_{1k} \\
\vdots & \ddots & \vdots \\
d_{k1} & \cdots & d_{kk}
\end{pmatrix} \in \mathbb{R}^{k \times k}.
\end{equation}

\begin{equation}
d_{ij} = x_{ij} + \boldsymbol{\alpha} \cdot q_{ij}
\end{equation}
The enhanced feature block will finally undergo a linear transformation through a learnable weight matrix \(\mathbf{W}\), and a bias \(b\) is added to obtain the final output \(\mathbf{Y}\):
\begin{equation}
\mathbf{Y} = \mathbf{W} \times \mathbf{D} + b = 
\sum_{i=1}^{k} \sum_{j=1}^{k} w_{ij} \left( x_{ij} + \boldsymbol{\alpha} \sum_{p=1}^{k} \sum_{q=1}^{k} x_{ij} x_{pq} \right) + b
\end{equation}

Wherein, the learnable weight matrix \(\mathbf{W}\) is:
\begin{equation}
\mathbf{W} = (w_{ij}) = 
\begin{pmatrix}
w_{11} & \cdots & w_{1k} \\
\vdots & \ddots & \vdots \\
w_{k1} & \cdots & w_{kk}
\end{pmatrix} \in \mathbb{R}^{k \times k}
\end{equation}

Through these three steps, the feature interaction mechanism introduces global interaction information between features while retaining the original features. Due to the application of the dendritic residual structure, its performance is not inferior to that of traditional convolution, and it can effectively improve the network’s ability to extract complex features.
\section{Experiment}
\subsection{Experimental Setup}
\subsubsection{Datasets}
To comprehensively verify the effectiveness of DDC in improving robustness in noisy environments, this study selects mainstream standard datasets and their perturbed variants in the field of computer vision, covering two core tasks: image classification and object detection. The specific details are as follows:

CIFAR-10 is a benchmark dataset for image classification tasks, containing 32×32 color images across 10 categories, with a total of 60,000 samples (50,000 for the training set and 10,000 for the test set). This dataset has a balanced sample distribution and universal scenarios, and can provide training data and the original test set for subsequent perturbation experiments.

To evaluate the anti-perturbation performance of classification tasks, this study adds 6 types of quantization noise perturbations to the CIFAR-10 test set to form test sets. Each noise type is configured within a specific parameter range: the normalized standard deviation of Gaussian noise is randomly selected from 0.01 to 0.12; the intensity coefficient of Poisson noise is randomly selected from 4 to 22; the pixel ratio of salt-and-pepper noise is randomly selected from 0.01 to 0.18; the noise intensity of speckle noise is randomly selected from 0.04 to 0.35; the scale parameter of Rayleigh noise is randomly selected from 0.08 to 0.45; the shape parameter of Gamma noise is randomly selected from 1.5 to 5.5, and the scale parameter is randomly selected from 0.08 to 0.22. The parameter ranges of the 6 selected quantization noises all refer to common disturbance intensity settings in image classification tasks and cover the range from slight to severe disturbances, which can effectively evaluate the model’s noise resistance at different disturbance levels.

To verify the generality of DDC in more complex visual tasks, the VOC2007 object detection dataset is selected. This dataset contains 20 object categories and approximately 9,000 images; 5,977 images are randomly selected as the training set, and the validation set and test set each contain 1,993 images. By referring to the disturbance settings of the classification task, the above 6 types of noise perturbations are added to the images in the VOC2007 test set. The detection precision $mAP@0.5$ and $mAP@0.5:0.95$ are compared to verify its noise resistance in object detection tasks.

\subsubsection{Evaluation Metrics}
For the classification task, $Accuracy$ is used as the primary metric to assess anti-perturbation robustness on perturbed test samples. The calculation formula is as follows:
\begin{equation}
\text{Accuracy} = \frac{TP + TN}{TP + TN + FP + FN} \times 100\%
\end{equation}

\begin{figure}[t!]
  \centering
  \subfloat[]{%
    \includegraphics[width=0.48\textwidth]{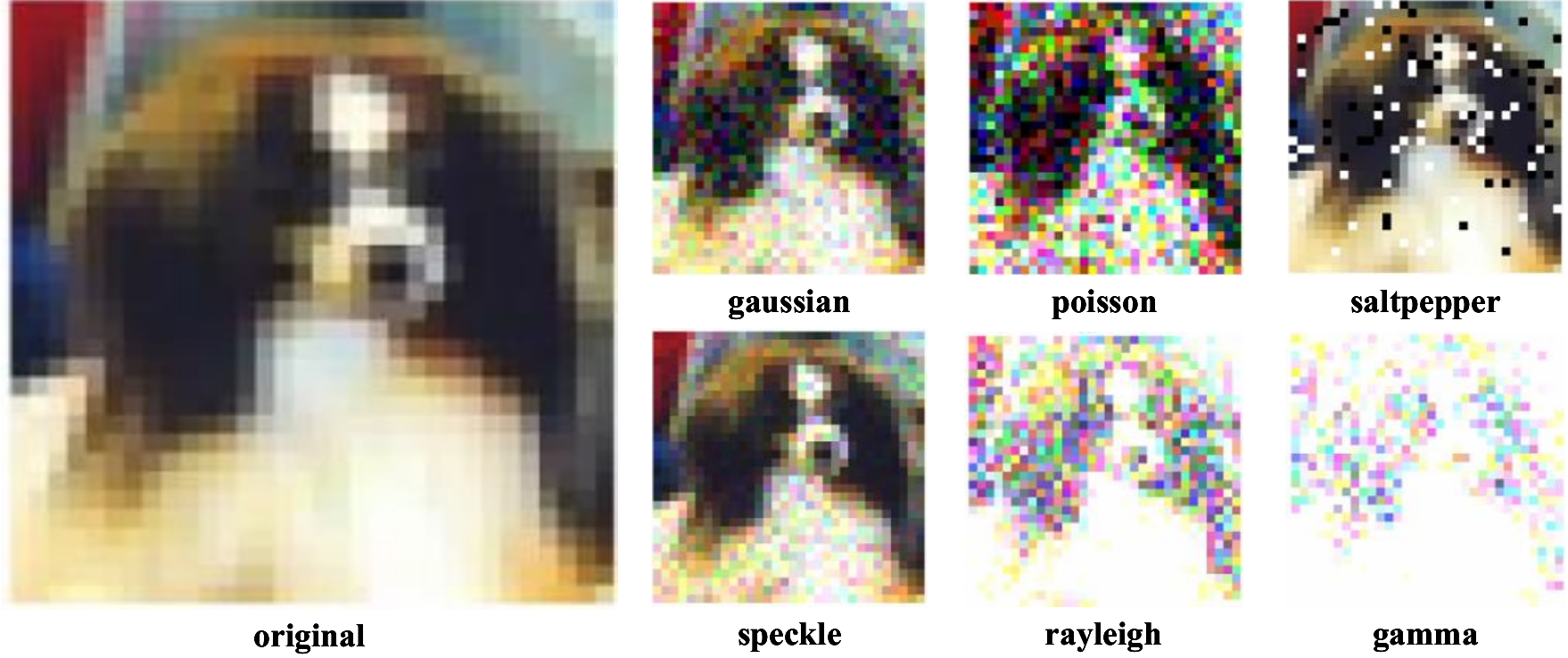}%
    \label{fig:suba}%
  }

  \vspace{8pt} 

  \subfloat[]{%
    \includegraphics[width=0.48\textwidth]{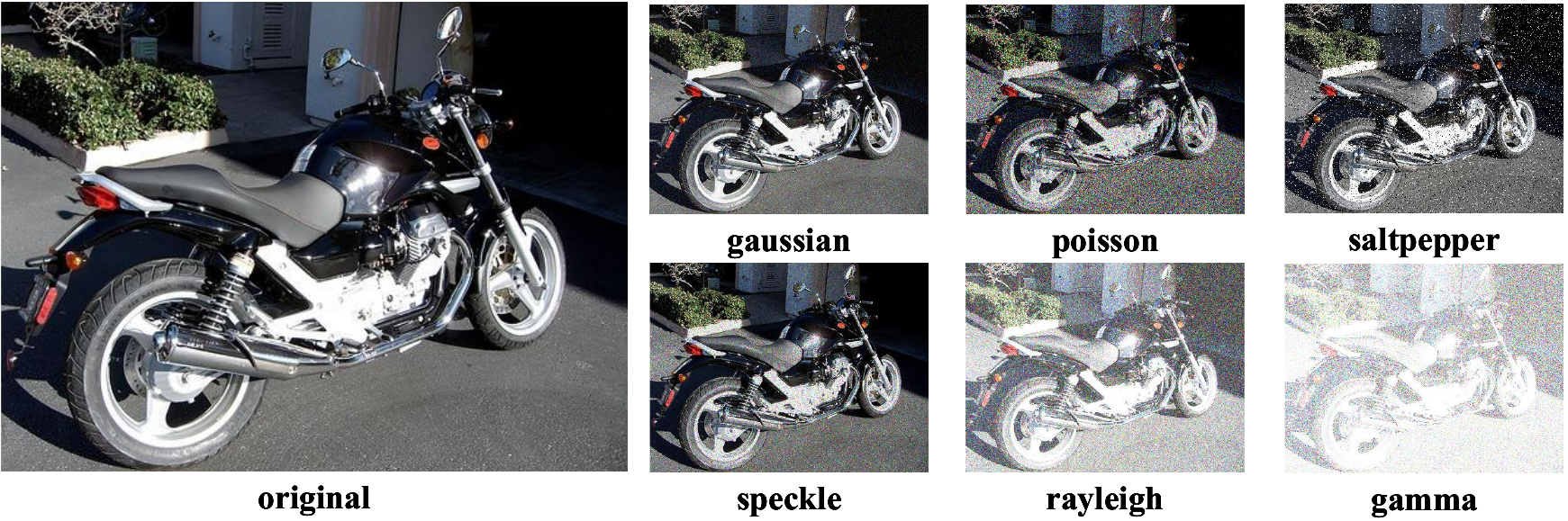}%
    \label{fig:subb}%
  }

  \caption{Visualization of Noisy Images; Figure (a) shows samples from the classification noise dataset of CIFAR-10, and Figure (b) shows samples from the detection noise dataset of VOC2007.}
  \label{fig:main}
\end{figure}
Among them,$TP$ (True Positive) refers to the number of samples that the model correctly predicts as positive;$TN$(True Negative) refers to the number of samples that the model correctly predicts as negative;$FP$(False Positive) refers to the number of samples that the model incorrectly predicts as positive; and $FN$(False Negative) refers to the number of samples that the model incorrectly predicts as negative.

For the object detection task, the mean Average Precision ($mAP$) is used as the main evaluation metric. By analyzing the changes in $mAP@0.5$ and $mAP@0.5:0.95$ on the VOC2007 noisy test set, the anti-disturbance capability of the model in complex tasks is quantified. The calculation formula is as follows:
\begin{equation}
\text{mAP@0.5} = \frac{1}{N} \sum_{c=1}^{N} AP@0.5_{c}
\end{equation}

Among them, $AP@0.5_{c}$ is the average precision for the c-th category, and its calculation formula is based on the integral of the precision-recall ($P-R$) curve of this category, that is:
\begin{equation}
\text{AP@0.5}_{c} = \sum_{i=1}^{m} \left( \text{Recall}_{i} - \text{Recall}_{i-1} \right) \times \max_{\text{Recall} \geq \text{Recall}_{i}} \text{Precision}
\end{equation}

Here, $m$ is the total number of detection boxes for this category sorted in descending order of confidence. $Recall_{i}$ is the recall rate of the first $i$ detection boxes, and $max_{Recall \geq Recall_{i}} Precision$ represents the maximum precision when the recall rate is not lower than $Recall_{i}$. The calculation formulas for recall rate and precision are as follows:
\begin{align}
\text{Recall} &= \frac{TP}{TP + FN} \label{eq:recall} \\
\text{Precision} &= \frac{TP}{TP + FP} \label{eq:precision}
\end{align}
\begin{table*}[t!]
  \centering
  \caption{Noise Resistance Performance Test (Image Classification)}
  \label{tab:noise_classification}
  
  \begin{tabular*}{\textwidth}{@{\extracolsep{\fill}}lccccccc@{}}
    \toprule
    Model & Convolution & \textbf{Gaussian} & \textbf{Poisson} & \textbf{Salt\&Pepper} & \textbf{Speckle} & \textbf{Rayleigh} & \textbf{Gamma} \\
    \midrule
    \multirow{2}{*}{YOLOv11-cls} 
    & Conv & 68.13\%$\pm$0.39\% & 22.14\%$\pm$2.11\% & 33.24\%$\pm$0.93\% &\bfseries57.22\%$\pm$0.72\% & 31.52\%$\pm$0.51\% & 19.39\%$\pm$0.92\% \\
    & DDC & \bfseries 68.56\%$\pm$0.54\% & \bfseries 24.62\%$\pm$0.48\% & \bfseries 33.66\%$\pm$1.31\% &  57.20\%$\pm$0.66\% & \bfseries 34.88\%$\pm$0.72\% & \bfseries 22.10\%$\pm$0.89\% \\
    \addlinespace
    \multirow{2}{*}{VGG16} 
    & Conv & 57.26\%$\pm$1.23\% & 14.98\%$\pm$0.24\% & 26.53\%$\pm$2.12\% & 49.56\%$\pm$1.21\% & 25.35\%$\pm$1.16\% & 18\%$\pm$0.63\% \\
    & DDC & \bfseries 60.66\%$\pm$1.79\% & \bfseries 18.36\%$\pm$2.9\% & \bfseries 30.15\%$\pm$2.35\% & \bfseries 52.18\%$\pm$2.41\% & \bfseries 27.01\%$\pm$1.46\% & \bfseries 18.32\%$\pm$0.7\% \\
    \addlinespace
    \multirow{2}{*}{EfficientNet-B0} 
    & Conv & 38.57\%$\pm$1.4\%  & 10.81\%$\pm$0.24\% & 33.75\%$\pm$3.57\% & 30.29\%$\pm$1.45\% & 15.97\%$\pm$0.44\% & 13.37\%$\pm$0.55\% \\
    & DDC & \bfseries 43.72\%$\pm$1.62\% & \bfseries 12.05\%$\pm$1.89\% & \bfseries 37.72\%$\pm$4.18\% & \bfseries 34.69\%$\pm$2.36\% & \bfseries 17.62\%$\pm$1.54\% & \bfseries 14.16\%$\pm$1.46\% \\
    \bottomrule
  \end{tabular*}
\end{table*}
In object detection, $TP$ refers to the number of prediction boxes that are successfully matched with any real box whose intersection over union ($IoU$) is greater than or equal to the threshold ( such as 0.5 ).$FP$ refers to the number of prediction boxes whose $IoU$ is lower than the threshold for all real boxes, or which are repeatedly matched with the same real box ;$FN$ refers to the number of real boxes that are not matched by any prediction box. The calculation formula for $mAP@0.5:0.95$ is as follows:
\begin{equation}
\text{mAP@0.5:0.95} = \frac{1}{10} \sum_{\text{IoU}=0.5}^{0.95,\ \text{step}=0.05} \text{mAP@IoU}
\end{equation}

That is, the $mAP$ under the 10 thresholds of $IoU$ from 0.5 to 0.95 ( step size 0.05 ) is averaged. The two types of indicators reflect the basic performance and robustness of the model under different strict ranges.

\subsubsection{Comparison Models}
To verify the generality and superiority of DDC across different types of backbone networks and tasks, this study selects three typical classification architectures in the field of computer vision (YOLOv11-cls, VGG16, and EfficientNet-B0) and three object detection architectures (YOLOv5, YOLOv8, and YOLOv11) as baselines. The original convolutional layers of each architecture are replaced with DDC, and the performance is tested on both the original dataset and the noise-added dataset.

\begin{figure*}[t!]
  \centering
  \includegraphics[width=1\linewidth]{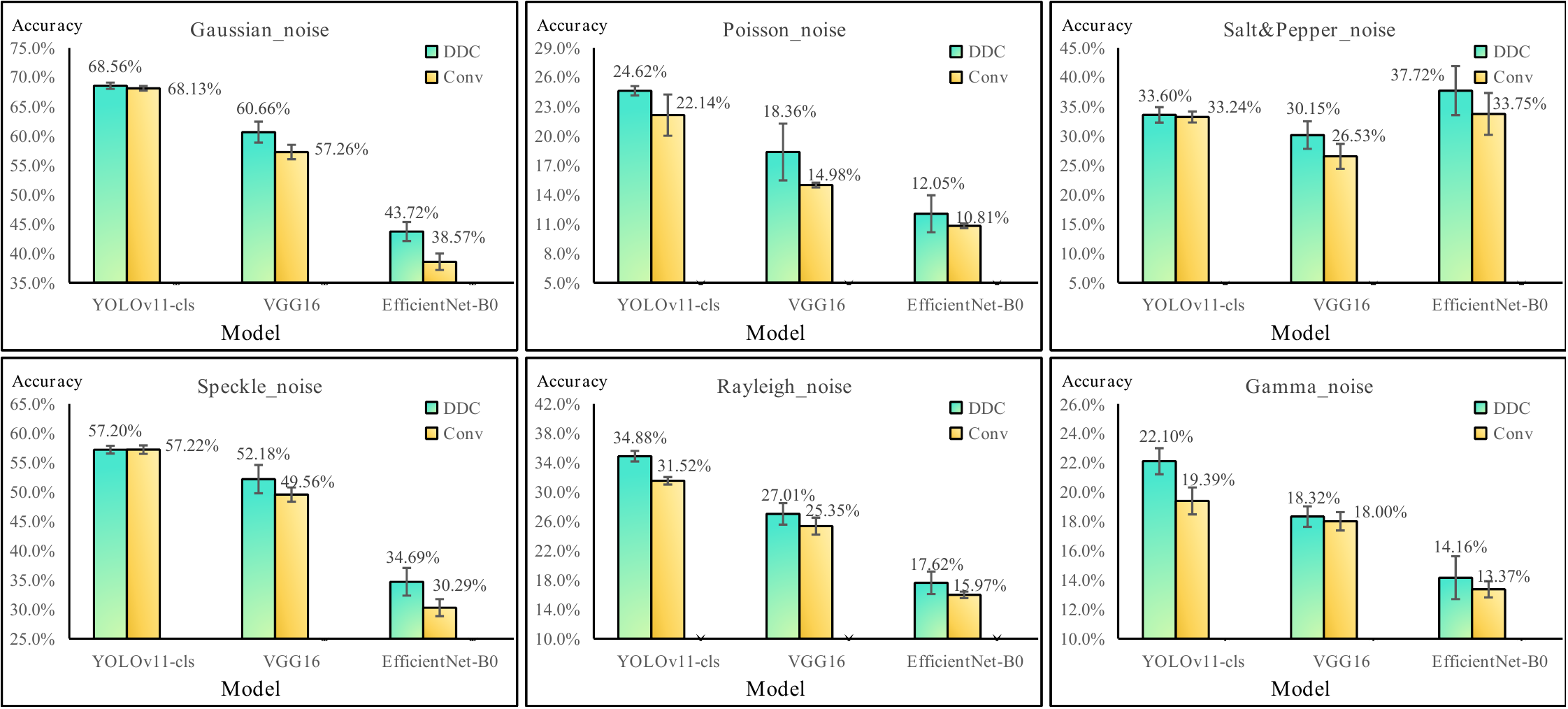} 
  \caption{Accuracy comparison of classification models under different noises. The abscissa represents the model type, and the ordinate represents the classification accuracy (\%). The yellow bar graph represents the performance of the model using traditional convolution ( Conv ), and the green bar graph represents the performance of the model using DDC. All data are the average of three test results. }
  \label{fig:comparison}
\end{figure*}
\begin{figure*}[t!]  
\centering
\includegraphics[width=1\linewidth]{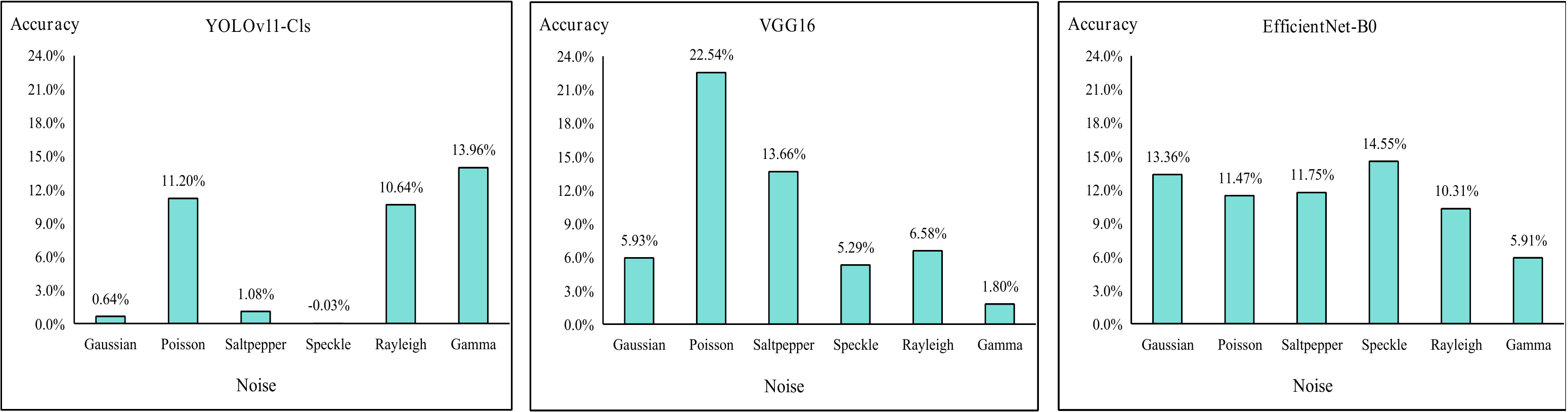}  
\caption{Percentage of Relative Improvement in Accuracy}
\label{fig:comparison}
\end{figure*}
\subsubsection{Parameter Settings}
All classification tasks in this experiment follow the same training settings: an initial learning rate of 0.001, a batch size of 64, and the SGD optimizer, with a total of 200 epochs trained. For object detection tasks, the training settings are as follows: an initial learning rate of 0.001, a batch size of 16, the Adam optimizer, and a total of 300 epochs trained. A random seed is used, and the hardware includes NVIDIA GeForce RTX 4080 GPU, 13th Gen Intel Core i9 CPU, and 32GB memory. The software environment includes Windows 11 operating system, Python 3.8.0, PyTorch 2.4.1, and the tcu121 framework. All hyperparameters, training strategies, and testing procedures are kept consistent to ensure the fairness of comparisons.

\subsection{Comparative Experiments}
To verify the anti-noise superiority of DDC in noise interference scenarios, \textbf{we conducted comparative experiments using three classification models and three detection models. For each type of model, two sets of architectures were designed: one retained the original convolutional layers, while the other replaced Conv with DDC while keeping all other components unchanged. }The performance differences between the two sets of architectures were systematically analyzed under six noise types (Gaussian noise, Poisson noise, SaltPepper noise, Speckle noise, Rayleigh noise, and Gamma noise) through two major tasks: image classification and object detection.

\subsubsection{Test on Anti-noise Performance of Image Classification}
On the noisy CIFAR-10 test set, three typical classification models, namely YOLOv11-cls, VGG16, and EfficientNet-B0, were selected to compare the accuracy differences between DDC and Conv under different noises. The results are shown in the following tables and charts.

It can be seen from TABLE II and Fig.5 that in addition to a slight decrease of 0.02\% in the performance of YOLOv11-cls under speckle noise, DDC achieves performance improvement in all models in the other five noise scenarios, which proves that it has stable anti-interference ability for most noise types. In addition, the performance improvement is quite significant. The performance improvement of EfficientNet-B0 model is the most obvious, and the overall average accuracy is improved by 4.05\%. VGG16 followed, with an average increase of 3.12\% ; the average increase in YOLOv11-cls was 2.21\%. This difference shows that DDC can bring more significant anti-interference gain for models with weak feature extraction ability. The reason is that these models are more sensitive to noise in the traditional convolution framework due to the lack of feature extraction ability, and the nonlinear interaction mechanism of DDC can better make up for their shortcomings in feature selection, so as to achieve more significant performance optimization. 

In addition, in Fig. 5, the relative accuracy improvement of EfficientNet-B0 under each interference is more stable, which further proves that DDC has a greater advantage in improving the anti-noise ability of such lightweight models. Models such as YOLOv11-cls, which have integrated anti-interference modules and strategies, have certain anti-noise capabilities. Therefore, under certain noises, the additional improvement brought by DDC is relatively small. In addition, according to Fig. 6, the relative increase percentage of EfficientNet-B0 and VGG16 under all six noises is positive, and only YOLOv11-cls has a slight decrease of 0.02\% under salt and pepper noise. These data further prove the adaptability and reliability of DDC for most noise scenarios and demonstrate its ability to replace traditional convolution.
\begin{table*}[t!]
  \centering
  \caption{Anti-Interference Performance Test (Object Detection) - mAP@0.5}
  \label{tab:anti_interference_detection_map50}
  \begin{tabular*}{\textwidth}{@{\extracolsep{\fill}}llcccccc@{}}
    \toprule
    \textbf {Model} & \textbf {Convolution} & \textbf{Gaussian} & \textbf{Poisson} & \textbf{Salt\&Pepper} & \textbf{Speckle} & \textbf{Rayleigh} & \textbf{Gamma} \\
    \midrule
    \multirow{2}{*}{YOLOv11} 
    & Conv & 0.4853$\pm$0.0173 & 0.1493$\pm$0.0151 & 0.2203$\pm$0.0205 & 0.4987$\pm$0.0201 & 0.1697$\pm$0.0126 & 0.0748$\pm$0.0056 \\
    & DDC & \bfseries 0.5051$\pm$0.0102 & \bfseries 0.1811$\pm$0.0258 & \bfseries 0.2494$\pm$0.0206 & \bfseries 0.5158$\pm$0.0084 & \bfseries 0.1923$\pm$0.0262 & \bfseries 0.0876$\pm$0.0213 \\
    \addlinespace
    \multirow{2}{*}{YOLOv8} 
    & Conv & 0.4894$\pm$0.0076 & 0.1456$\pm$0.0140 & 0.2262$\pm$0.0115 & 0.4940$\pm$0.0049 & 0.1748$\pm$0.0042 & 0.0708$\pm$0.0024 \\
    & DDC & \bfseries 0.5229$\pm$0.0049 & \bfseries 0.1954$\pm$0.0063 & \bfseries 0.2678$\pm$0.0155 & \bfseries 0.5236$\pm$0.0082 & \bfseries 0.2105$\pm$0.0155 & \bfseries 0.0942$\pm$0.0148 \\
    \addlinespace
    \multirow{2}{*}{YOLOv5} 
    & Conv & \bfseries 0.4860$\pm$0.0010 & 0.1525$\pm$0.0153 & 0.2163$\pm$0.0094 & \bfseries 0.4891$\pm$0.0062 & 0.1759$\pm$0.0077 & 0.0734$\pm$0.0123 \\
    & DDC &  0.4846$\pm$0.0024 & \bfseries 0.1633$\pm$0.0063 & \bfseries 0.2366$\pm$0.0274 &  0.4880$\pm$0.0067 & \bfseries 0.1863$\pm$0.0095 & \bfseries 0.0869$\pm$0.0110 \\
    \bottomrule
  \end{tabular*}
\end{table*}

\begin{table*}[t!]
  \centering
  \caption{Anti-Interference Performance Test (Object Detection) - mAP@0.5:0.95}
  \label{tab:anti_interference_detection_map50_95}
  \begin{tabular*}{\textwidth}{@{\extracolsep{\fill}}llcccccc@{}}
    \toprule
    \textbf {Model} & \textbf {Convolution} & \textbf{Gaussian} & \textbf{Poisson} & \textbf{Saltpepper} & \textbf{Speckle} & \textbf{Rayleigh} & \textbf{Gamma} \\
    \midrule
    \multirow{2}{*}{YOLOv11} 
    & Conv & 0.3170$\pm$0.0147 & 0.0872$\pm$0.0086 & 0.1358$\pm$0.0127 & 0.3255$\pm$0.0148 & 0.1061$\pm$0.0083 & 0.0436$\pm$0.0027 \\
    & DDC & \bfseries 0.3338$\pm$0.0124 & \bfseries 0.1074$\pm$0.0166 & \bfseries 0.1568$\pm$0.0151 & \bfseries 0.3372$\pm$0.0115 & \bfseries 0.1242$\pm$0.0190 & \bfseries 0.0534$\pm$0.0159 \\
    \addlinespace
    \multirow{2}{*}{YOLOv8} 
    & Conv & 0.3162$\pm$0.0051 & 0.0863$\pm$0.0080 & 0.1388$\pm$0.0059 & 0.3217$\pm$0.0038 & 0.1089$\pm$0.0028 & 0.0422$\pm$0.0006 \\
    & DDC & \bfseries 0.3421$\pm$0.0024 & \bfseries 0.1190$\pm$0.0061 & \bfseries 0.1665$\pm$0.0132 & \bfseries 0.3449$\pm$0.0071 & \bfseries 0.1361$\pm$0.0148 & \bfseries 0.0582$\pm$0.0118 \\
    \addlinespace
    \multirow{2}{*}{YOLOv5} 
    & Conv & \bfseries  0.3060$\pm$0.0050 & 0.0877$\pm$0.0108 & 0.1295$\pm$0.0085 & \bfseries 0.3120$\pm$0.0076 & 0.1098$\pm$0.0078 & 0.0450$\pm$0.0105 \\
    & DDC & 0.2999$\pm$0.0028 & \bfseries 0.0925$\pm$0.0063 & \bfseries 0.1375$\pm$0.0174 &  0.3001$\pm$0.0028 & \bfseries 0.1106$\pm$0.0092 & \bfseries 0.0497$\pm$0.0086 \\
    \bottomrule
  \end{tabular*}
\end{table*}
\begin{figure*}[t!]
  \centering
  \subfloat[]{%
    \includegraphics[width=1\textwidth]{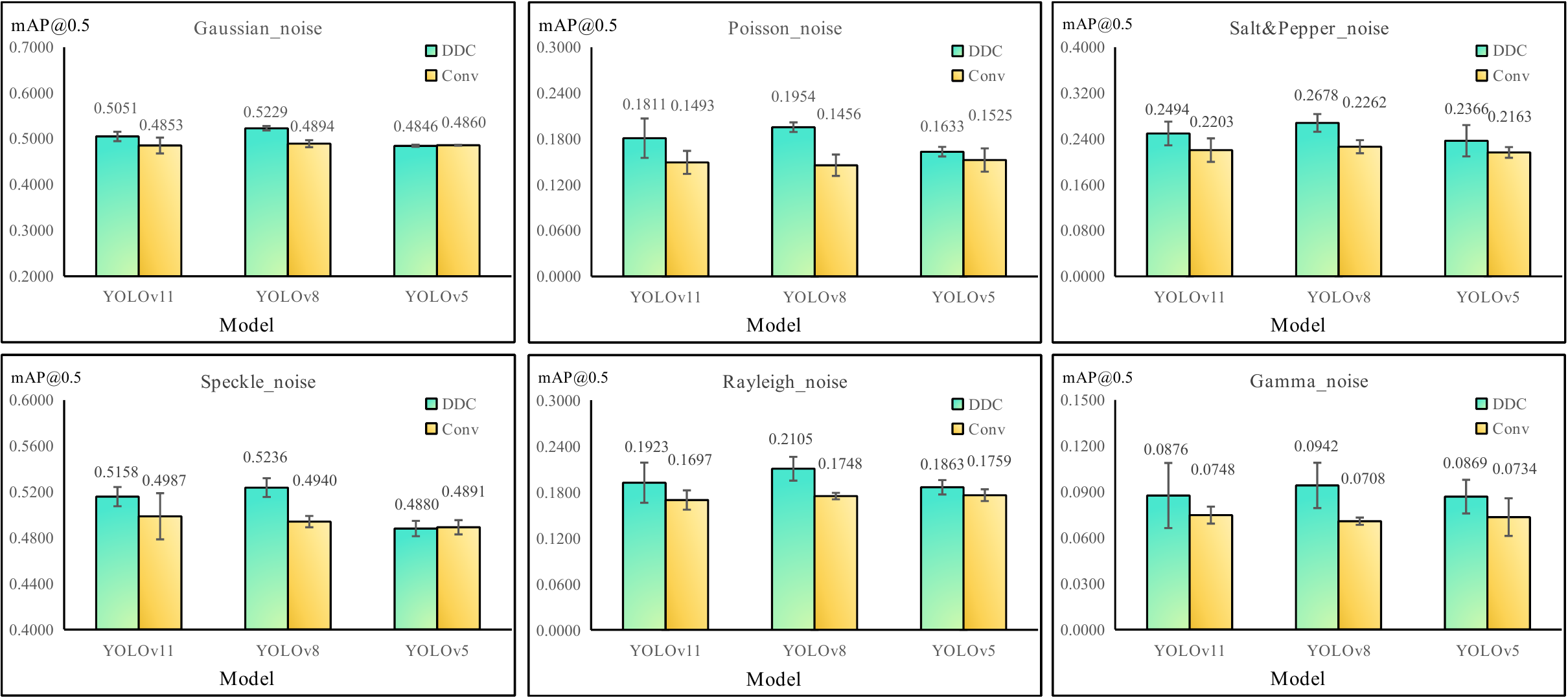}%
    \label{fig:suba}%
  }

  \vspace{2pt} 

  \subfloat[]{%
    \includegraphics[width=1\textwidth]{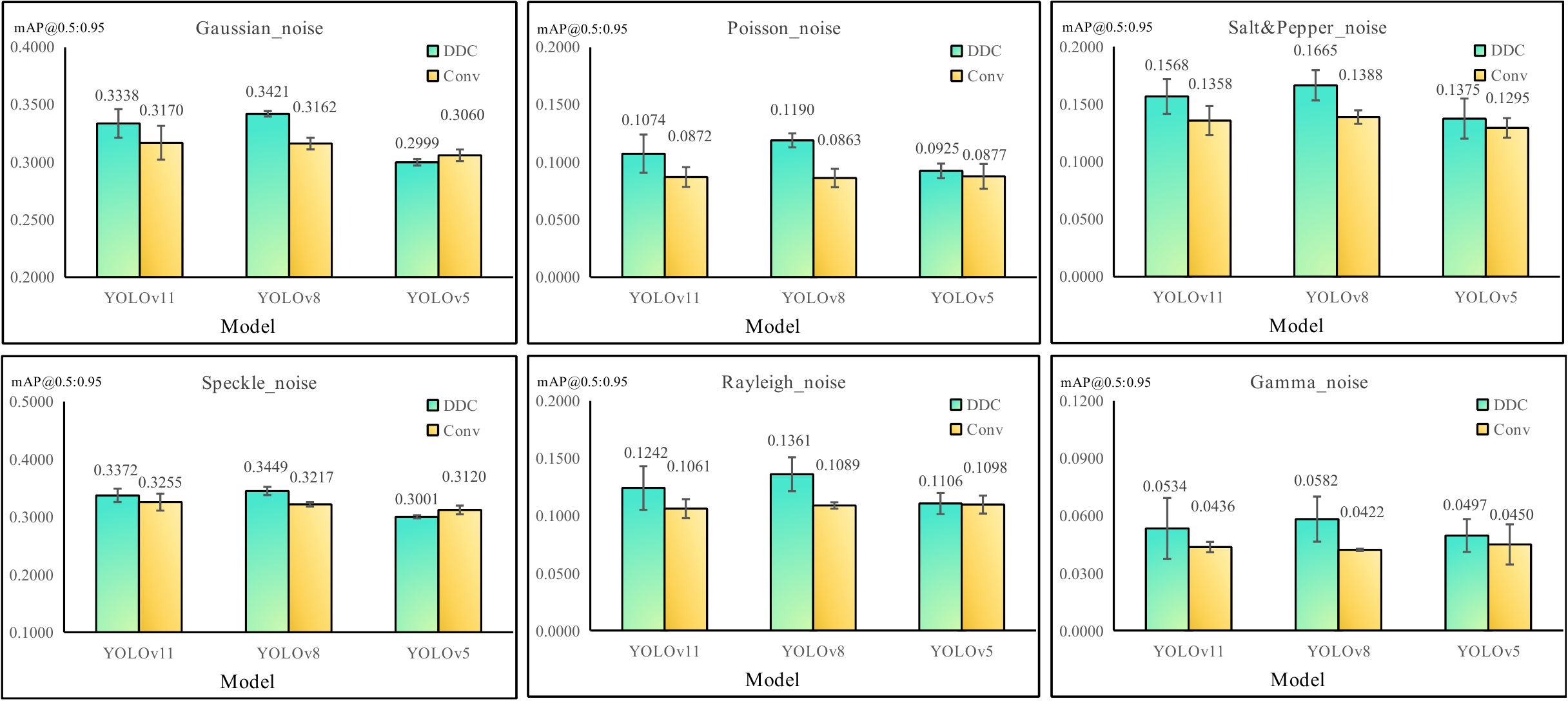}%
    \label{fig:subb}%
  }

  \caption{Bar Chart Comparison of $mAP@0.5$ (a) and $mAP@0.5:0.95$ (b) of Detection Models under Different Noises}
  \label{fig:main}
\end{figure*}

\subsubsection{ Anti-noise Performance Evaluation for Object Detection}
On the noisy test set of VOC2007, three detection models, namely YOLOV11, YOLOV8, 
and YOLOv5, are selected. With $mAP@0.5$ and $mAP@0.5:0.95$ as the evaluation criteria, the performance of DDC and Conv is compared, and the results are shown in the TABLE III.

It can be seen from  TABLE III and Fig. 7 that in the object detection task, DDC has different performance in improving the anti-interference performance of different models, and has stable noise adaptability as a whole. Among them, YOLOv8 performs best. After replacing the traditional convolution with DDC, its $mAP@0.5$ and mAP@0.5: 0.95 are significantly improved in six noise scenarios, and the improvement is the largest in the three models. For example, under Gaussian noise, they are increased by about 6.8\% and 8.2\% respectively, and under salt and pepper noise, they are increased by about 18.3\% and 20.0\% respectively. This is because the network architecture of YOLOv8 is highly compatible with the nonlinear interaction mechanism of DDC. Its feature extraction process can make full use of the neighborhood correlation information generated by DDC, and reduce the direct impact of noise by shifting from focusing on a single pixel signal to focusing on the interaction information between neighborhood features, thereby releasing greater performance potential in complex noise environments. YOLOv11 also shows robust and comprehensive performance improvement. The data show that the $mAP@0.5$ and mAP@0.5: 0.95 indicators have achieved steady growth under six kinds of noise interference. Especially in Poisson noise and gamma noise scenarios, the performance improvement is particularly significant, $mAP@0.5$ is increased by about 21.25\% and 17.12\%, respectively. This further confirms that the DDC module effectively enhances the robustness of the YOLOv11 architecture.

For YOLOv5, its overall performance also verifies the advantages of DDC. Although the $mAP@0.5$ decreases by 0.27\% and 0.22\% respectively under Gaussian noise and speckle noise, such fluctuations are essentially optimization problems at the technical implementation level, rather than the defects of the DDC mechanism itself. It can be seen from Fig. 7 that in the six noise tests, the $mAP@0.5$ of YOLOv5 increases by more than 10\% under the four noises of Poisson, salt and pepper, Rayleigh and gamma, and the gain is significant. In contrast, a small decrease of less than 0.3\% under Gaussian and speckle noise can be regarded as a normal fluctuation in the overall trend. Although there is a slight performance fluctuation under individual noise types, it does not affect its overall advantage. Therefore, this slight fluctuation will not weaken the effectiveness of DDC as a basic anti-noise technology, and such problems can be completely solved in the future through targeted compatibility parameter tuning ( such as adjusting the balance factor $\boldsymbol{\alpha}$ ).

In Poisson noise and Gamma noise scenarios, DDC achieves more prominent optimization effects. Taking Gamma noise as an example: YOLOv8’s $mAP@0.5$ and $mAP@0.5:0.95$ are relatively improved by 33.00\% and 38.02\% respectively; YOLOv11’s corresponding indicators increase by 17.32\% and 22.31\% respectively; YOLOv5’s see relative gains of 35.10\% and 33.27\% respectively. These quantitative results demonstrate that DDC delivers superior performance enhancement in signal-dependent noise environments compared to other noise types.

In summary, in the experiments of image classification and object detection, replacing traditional convolution with DDC is a general strategy to effectively enhance the anti-noise robustness of the model. Whether in relatively simple image classification tasks or more complex object detection scenarios, DDC can inject powerful noise filtering capabilities into the model through its bionic dendritic computing paradigm.
\section{Discussion}
Inspired by the dendritic information processing mechanism of biological neurons, this study proposes Dendritic Convolution. It breaks through the robustness bottleneck of traditional convolution in noisy environment by innovating at the bottom of feature extraction logic. \textcolor{Magenta}{\textbf{The fundamental reason why DDC can effectively improve the robustness of the model in complex noise environment is the introduction of neighborhood feature interaction mechanism. In traditional convolution, noise points are directly involved in the weighted calculation, resulting in noise interference directly propagating to the feature representation. In contrast, by simulating the nonlinear preprocessing function of neuron dendrites, DDC can transform the logic of feature extraction from the independent signal dependent on a single pixel to the correlation analysis and integration of multiple pixel signals in the neighborhood, which will naturally highlight the information that is highly correlated with the whole neighborhood. When the noise point is identified, it will be naturally weakened due to its low correlation with the surrounding effective pixels, and will not dominate the feature output, thereby reducing the direct impact of noise ( see Fig. 8 ), so that DDC can complete the initial noise filtering in the feature extraction stage, which also enables it to achieve more significant performance improvement in signal-dependent noise scenarios.}}
\begin{figure}[t]
  \centering
  \includegraphics[width=1\linewidth]{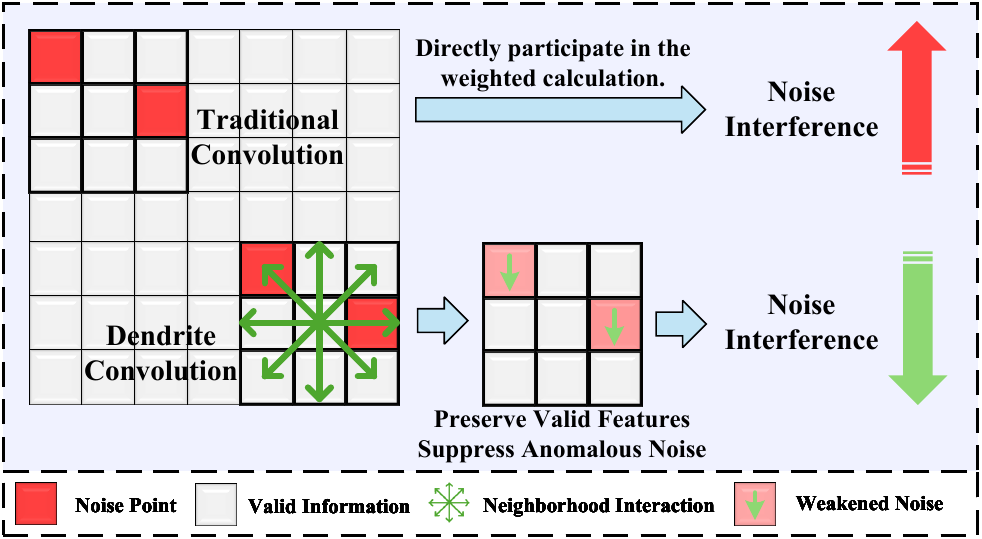} 
  \caption{Schematic Diagram of Noise Suppression Mechanism in Dendritic Convolution}
  \label{fig:comparison}
\end{figure}
In addition, Dendritic Convolution can be directly integrated into different types of CNN frameworks. Even for lightweight models with relatively weak feature extraction ability, DDC can make up for their shortcomings in complex feature screening ability and significantly enhance noise robustness. Future work will focus on the adaptability between different parameters and DDC, explore lightweight improvement schemes for specific architectures, and optimize computational efficiency while maintaining performance advantages.

\section{Conclusion}
To address the complex noise interference challenge faced by image recognition models in real-world scenarios, this paper proposes a biologically inspired Dendritic Convolution. Breaking through the limitation of traditional convolution’s linear weighted summation, this method simulates the nonlinear signal integration mechanism of biological neuronal dendrites and introduces interaction terms between local features into the convolution operation, thereby realizing the preprocessing of input features and enhancing the model’s adaptive noise filtering capability. Validation results on the CIFAR-10 and PASCAL VOC2007 datasets for image classification and object detection tasks show that, without relying on additional network modules or complex data preprocessing strategies, DDC can significantly improve the robustness of different types of backbone networks under various noise interferences only by replacing traditional convolution, while exhibiting excellent universality.

\newpage

\end{document}